\documentclass{ifacconf}
\usepackage{graphicx}      
\usepackage{natbib} 
\usepackage{algorithmic}
\usepackage{algorithm}

\usepackage{pgfplots}
\usepackage{graphics} 
\usepackage{epsfig}
\usepackage{amsmath,mathtools} 
\usepackage{empheq}
\usepackage{amssymb, psfrag, booktabs}
\usepackage{xcolor}
\usepackage{bytefield}
\usepackage{verbatim}
\usepackage{fmtcount}
\usepackage{siunitx}
\usepackage{multirow}
\usepackage{pstricks}
\usepackage[flushleft]{threeparttable}
\usetikzlibrary{shapes.geometric, arrows.meta, positioning}
\usepackage[hyphens]{url}

\makeatletter
\let\AND\original@AND
\makeatother

\newcommand{\ur}{UR10\lowercase{e}}

\usetikzlibrary{positioning, shapes, calc, fit, backgrounds}
\usepackage[dvipsnames]{xcolor}
\usepackage[svgnames]{xcolor}
\usepackage[x11names]{xcolor}
\begin{document}
\begin{frontmatter}		
\title{Iterative Tuning of Nonlinear Model Predictive Control for Robotic Manufacturing Tasks \thanksref{footnoteinfo}} 
\thanks[footnoteinfo]{This work was supported as a part of NCCR Automation, a National Centre of 
Competence in Research, funded by the Swiss National Science Foundation (grant number 
51NF40\textunderscore225155), and partially by Joh. Jakob Rieter Stiftung and Rieter AG.}
\author[First]{Deepak Ingole} 
\author[First]{Valentin Bhend}
\author[First]{Shiva Ganesh Murali}
\author[Second]{Oliver Döbrich}
\author[First]{Alisa Rupenyan}
\address[First]{ZHAW Centre for Artificial Intelligence, \\Zürich University of Applied Sciences, \\
Winterthur, 8400, Switzerland,\\(e-mail:\{inge, rupn\}@zhaw.zh, 
\{bhendval, muralshi\}@students.zhaw.ch).}
\address[Second]{ZHAW Institute of Materials and Process Engineering \\Zürich University of 
Applied Sciences, \\ Winterthur, 8400, Switzerland, \\(e-mail: doec@zhaw.ch).}
\begin{abstract} 		
Manufacturing processes are often perturbed by drifts in the environment and wear in the system, requiring control re-tuning even in the presence of repetitive operations. This paper presents an iterative learning framework for automatic tuning of Nonlinear Model Predictive Control (NMPC) weighting matrices based on task-level performance feedback. Inspired by norm-optimal Iterative Learning Control (ILC), the proposed method adaptively adjusts NMPC weights $Q$ and $R$ across task repetitions to minimize key performance indicators (KPIs) related to tracking accuracy, control effort, and saturation. Unlike gradient-based approaches that require differentiating through the NMPC solver, we construct an empirical sensitivity matrix, enabling structured weight updates without analytic derivatives. The framework is validated through simulation on a \ur~robot performing carbon fiber winding on a tetrahedral core. Results demonstrate that the proposed approach converges to near-optimal tracking performance (RMSE within 0.3\% of offline Bayesian Optimization (BO)) in just 4 online repetitions, compared to 100 offline evaluations required by BO algorithm. The method offers a practical solution for adaptive NMPC tuning in repetitive robotic tasks, combining the precision of carefully optimized controllers with the flexibility of online adaptation.
\end{abstract}		
\end{frontmatter}
\section{Introduction}
\label{sec:intro}
Advanced manufacturing increasingly relies on intelligent and sustainable automation, where robotic systems must adapt to variability in processes, materials, and task conditions. In many industrial settings, robots repeatedly execute similar motions across multiple cycles, making iterative tasks a natural domain for learning-based control improvement~\citep{alatartsev2015robotic}. To achieve high performance with minimal manual engineering effort, there is a growing interest in control frameworks that can automatically tune their parameters using data from repeated executions.

Nonlinear Model Predictive Control (NMPC) has become a leading paradigm for constrained motion control of robotic manipulators due to its ability to incorporate nonlinear dynamics, actuator limits, and task-space objectives~\citep{aro2024robust}. However, practical NMPC implementations typically require expert tuning of weighting matrices, prediction horizon, and regularization parameters. Manual tuning is time-consuming, system-specific, and often leads to suboptimal performance when task conditions change. Consequently, automated and data-driven NMPC tuning has emerged as an important research direction~\citep{allamaa2024learning}.

A wide range of NMPC tuning strategies has been explored to automatically adjust controller parameters under varying operating conditions. Several works focus on weight tuning, including Bayesian Optimization (BO)–based methods that optimize cost weights, penalties, or hyperparameters in a sample-efficient manner~\citep{roveda2019control, rupenyan2021performance,almasalmah2024auto}, adversarial-objective–based real-time weight adaptation for nonlinear MPC~\citep{ishihara2025real}, and optimal weight adaptation for connected and automated vehicles using Bayesian optimization. Neural Network–based approaches similarly adjust NMPC weights online to enhance tracking robustness under varying conditions~\citep{elsisi2021effective}, while Reinforcement Learning–driven schemes tune weights through reward-guided closed-loop interaction~\citep{salaje2024learning}. 

Beyond learning-based weight tuning, Linear Quadratic Regulator (LQR)-based weight transformations and economic tracking cost synthesis tools such as TuneMPC provide systematic cost construction for NMPC~\citep{de2020tunempc}. Broader developments include survey-based tuning guidelines~\citep{alhajeri2020tuning} and deep-learning–based self-tuning controllers~\citep{mehndiratta2021can}. Parallel to weight tuning, prediction-horizon tuning has also been studied, with variable-horizon self-tuning NMPC methods improving efficiency and responsiveness in nonlinear motion systems~\citep{wei2021variable}. 

Although the aforementioned methods represent meaningful progress in automated NMPC weight and horizon adaptation, the existing approaches generally rely on offline or single-shot optimization procedures, depend on large training datasets or computationally intensive learning processes, and may compromise NMPC’s inherent constraint-handling guarantees. Furthermore, both learning-based and optimization-based schemes typically treat each control episode independently and therefore fail to exploit the structured repetition and repetition-to-repetition tracking information that naturally arise in many robotic applications. These limitations underscore the need for a principled self-tuning NMPC framework that leverages repetitive task execution and closed-loop performance data across repetitions - a gap that is directly addressed by the Iterative Learning Control (ILC) inspired approach proposed in this work. 

Standard ILC provides a complementary mechanism for improving performance in iterative robotic tasks by refining control-related quantities based on previous-cycle tracking errors~\citep{bristow2006survey, barton2010norm}. ILC has been widely applied to enhance feedforward compensation, improve precision motion systems, and refine robot trajectory tracking through repetition~\cite{balta2021learning, wu2026autogeneration}. Beyond trajectory refinement, ILC has also been explored for controller tuning itself, where iterative updates adjust feedback gains or controller parameters across repetitions~\cite{lee2022ilcgains}. Related iterative learning–based tuning approaches include automatic Proportional–Integral–Derivative (PID) tuning and fixed-structure controller design using Linear Matrix Inequalities (LMI)-based ILC laws for stochastic systems~\citep{lee2022ilcgains}. Despite these developments, the integration of ILC and related methods with NMPC for automatic tuning of NMPC cost weights remains largely	under-explored.

We thus propose a self-tuning NMPC framework that leverages an iterative tuning approach to automatically adapt NMPC weighting matrices ($Q, R$) based on repetition-to-repetition tracking errors. The contribution is a general control architecture for iterative robotic tasks, where NMPC governs intra-repetition optimal control, while the iterative tuning procedure performs inter-repetition adaptation of cost parameters. This separation preserves NMPC’s constraint-handling and dynamic-optimization capabilities while enabling rapid, data-driven performance improvement without manual tuning. The contributions of this paper are:

\begin{itemize}
\item \textit{Iterative tuning NMPC architecture:} A general control framework that separates intra-repetition optimal control (NMPC) from inter-repetition parameter adaptation (iterative tuning procedure). The weighting matrices $\mathbf{Q}$ and $\mathbf{R}$ are parameterized and updated through a simplified norm-optimal ILC-inspired approach to minimize task-level KPIs.
\item \textit{Industrial case study - autonomous carbon-fiber winding:} An end-to-end autonomous 	winding system for a \ur~robot that integrates NMPC-based path planning and trajectory tracking with ILC-inspired weight adaptation. The repetitive multi-layer winding process on non-axisymmetric core provides an ideal testbed for the proposed approach.
\item \textit{Simulation validation:} Comprehensive evaluation on complex 3D geometry (tetrahedron core) demonstrating rapid convergence and significant tracking accuracy improvement over fixed-weight NMPC within few repetitions.
\item \textit{Comparison:} We compared a proposed iterative tuning approach with the Bayesian Optimization-based weight identification.
\end{itemize}
\section{Nonlinear Model Predictive Control}
\label{sec:nmpc}
Nonlinear MPC is a feedback control strategy for constrained nonlinear multi-variable systems. At each sampling time, NMPC solves a Constrained Finite-Time Optimal Control (CFTOC) problem by computing an optimal control sequence over a prediction horizon based on the current system state. Only the first control input is applied, and the optimization is repeated at the next sampling instant with updated measurements. This Receding Horizon Control (RHC) approach provides inherent feedback, enabling robustness against modeling errors and disturbances~\citep[Chapter 13]{borrelli2017predictive}.

We consider the following discrete-time nonlinear dynamic model of the plant for predictions
\begin{subequations}
\label{eq:nmpc:dlti}
\begin{align}
x_{k+T_s} &= f(x_k, u_k), \label{eq:nmpc:dlti:state}\\
y_k & = g(x_k,u_k),\label{eq:nmpc:dlti:output}
\end{align}
\end{subequations}
where $x\in \mathbb{R}^{n_x}$, $y\in \mathbb{R}^{n_y}$, and $u\in \mathbb{R}^{n_u}$ are the vector of state, output, and input, respectively. $T_s$ is the system sampling time and $k$ is the time step. 

\subsection{Problem Formulation}
\label{subsec:nmpc:form}
Using the nonlinear model~\eqref{eq:nmpc:dlti}, the nonlinear MPC problem could be formulated as a CFTOC problem for reference tracking as follows:
\begin{subequations}
\label{eq:cftocp:nmpc}	
\begin{align}
\label{eq:cftocp:nmpc:model:cost}
\min _{U}\sum_{k=0}^{N-1} & \frac{1}{2}\left(\vert\vert y_k-y^{\textrm{ref}}_k\vert\vert 
^{2}_{Q}+\vert\vert u_k-u_{k-1}\vert\vert^{2}_{R}\right)
\end{align}
\vspace{-3mm}
\begin{align}
\nonumber
&\text{s.t.}\;\;\\
\label{eq:cftocp:nmpc:model:states}
&x_{k+T_s}=f(x_k,u_k),& \forall k \in \{0,\dots,N-1\},\\ 
\label{eq:cftocp:nmpc:model:output}
&y_{k} = g(x_k,u_k), & \forall k \in \{0,\dots,N-1\},\\ 
\label{eq:cftocp:nmpc:const:states}
&x_{\min} \le x_k \le x_{\max},& \forall k \in \{0,\dots,N-1\},\\ 
\label{eq:cftocp:nmpc:const:input}
&u_{\min} \le u_k \le u_{\max}, & \forall k \in \{0,\dots,N-1\},\\ 
\label{eq:cftocp:nmpc:const:output}
&y_{\min} \le y_k \le x_{\max},& \forall k \in \{0,\dots,N-1\},\\ 
\label{eq:cftocp:nmpc:const:tstates}
&x_{\rm N, \min} \le x_N \le x_{\rm N,\max},&
\end{align}
\end{subequations}
where the term \eqref{eq:cftocp:nmpc:model:cost} is the objective function with prediction horizon $N$ which is weighted by $Q \succeq 0$ and $R \succ 0$, $T_s$ is the sampling time and the optimization is performed with respect to sequence of predicted optimal inputs $U_0 ={ u_0,\dots,u_{N-1}}$. 
By solving~\eqref{eq:cftocp:nmpc} with a given initial condition $u_{-1}= u(t-T_s)$ and $x_{0}= x(t)$ the optimization yields the optimal input sequence $U_0^{\star} = \{u_0^{\star}, \ldots, u_{N-1}^{\star}\}$ from which only the first element, i.e., $u_0^{\star}$ is implemented to the plant.
\vspace{-4.5mm}
\subsection{Problem Statement}
\label{subsec:problem}
Performance of NMPC depends critically on the choice of output and input weighting matrices $\mathbf{Q}$ and $\mathbf{R}$ in the cost function~\eqref{eq:cftocp:nmpc:model:cost}, which determine the trade-off between tracking accuracy, control smoothness, and actuator utilization. In nonlinear robotic systems, these weights cannot be selected systematically through analytical means, and small changes in task geometry or operating conditions often require substantial re-tuning~\citep{garriga2010model, alhajeri2020tuning}. This motivates an \emph{outer-loop learning mechanism} that automatically adjusts NMPC weights across task repetitions.

In repetitive robotic processes, e.g., pick-and-place, welding, assembly, fiber placement, and composite manufacturing, the same trajectory is executed multiple times with slight variations in geometry, payload, or environmental conditions. In such applications, the error metrics (overshoot at trajectory bends, velocity ripple, or prolonged actuator saturation) are very similar in every repetition, making them excellent candidates for data-driven, repetition-to-repetition learning~\citep{wu2026autogeneration}. While NMPC provides an effective framework for handling constraints and nonlinear dynamics, fixed weighting parameters present several challenges in iterative robotic applications, in particular when handling task variability in production. An efficient self-tuning procedure would be beneficial when switching from one task to another. Moreover, within-repetition tuning would take care of disturbances from material inconsistencies, tool wear, or thermal effects during operation on a single manufacturing task.	

\section{Iterative NMPC Weight Tuning}
\label{sec:ilc}
To automatically tune the NMPC weighting matrices $\mathbf{Q}$ and $\mathbf{R}$, we embed an algorithm inspired from \emph{Norm-Optimal ILC} (NO-ILC) in an outer loop that operates across task repetitions. At each repetition $\ell$, the robot executes the complete task using the current weight vector ${W}_\ell = [\text{vec}({Q}_\ell )^\top, \text{vec}({R}_\ell )^\top]^\top$ with element-wise bounds $Q_{\ell} \in [Q_{\min}, Q_{\max}],\;R_{\ell} \in [R_{\min}, R_{\max}]$, and upon completion we evaluate a set of task-level Key Performance Indicators (KPIs) that capture critical aspects of execution quality. At the end of each repetition $\ell$, each KPI is compared against its desired target value, producing a normalized error vector $e_\ell$, defined as:
\begin{equation}
e_\ell
= 
\begin{bmatrix}
e_\ell^{\mathrm{RMSE}} \\
e_\ell^{\Delta u} \\
e_\ell^{\mathrm{sat}} \\
e_\ell^{\max\mathrm{EE}}
\end{bmatrix}
=
\begin{bmatrix}
\displaystyle
\frac{\mathrm{RMSE}_\ell- \mathrm{RMSE}^{\star}}
{\mathrm{RMSE}^{\star}}
\\
\displaystyle
\frac{\mathrm{RMS}_{\Delta u,\ell} - \mathrm{RMS}_{\Delta u}^{\star}}
{\mathrm{RMS}_{\Delta u}^{\star}}
\\
\displaystyle
\frac{\mathrm{sat\_ratio}_\ell - \mathrm{sat}^{\star}}
{\mathrm{sat}^{\star}}
\\
\displaystyle
\frac{\max\mathrm{EE}_\ell - \max\mathrm{EE}^{\star}}
{\max\mathrm{EE}^{\star}}
\end{bmatrix},
\label{eq:ek:definition}
\end{equation}
where:
\begin{itemize}
\item $\mathrm{RMSE}_\ell$ is the output tracking Root Mean Square Error (RMSE) for repetition 
$\ell$,
\item $\mathrm{RMS}_{\Delta u,\ell}$ is the control-rate RMS (smoothness metric),
\item $\mathrm{sat\_ratio}_\ell$ is the fraction of time steps where actuators
reach torque or velocity saturation,
\item $\max\mathrm{EE}_\ell$ is the maximum instantaneous tracking error,
\item $\mathrm{RMSE}^{\star}$, $\mathrm{RMS}_{\Delta u}^{\star}$,
$\mathrm{sat}^{\star}$, and $\max\mathrm{EE}^{\star}$ are
user-defined target values.
\end{itemize}
A value $e_\ell > 0$ indicates underperformance with respect to the target, while $e_\ell < 0$ indicates better-than-target behavior. The goal of the outer-loop iterative tuning is to adjust the NMPC weight vector $W_\ell$ such that $e_\ell \rightarrow {0}$ as $\ell$ increases. 
\subsection{Simplified Norm-Optimal Iterative Tuning Law}
\label{subsec:noilc}
The iterative tuning formulation assumes a local affine relationship between changes in NMPC weights and changes in the KPI vector:
\begin{equation}
\label{eq:ilc_affine}
{e}_{\ell +1} = {e}_\ell  + {S}  \Delta {W},
\end{equation}
where ${S}\in {R}^{n_e \times n_w}$ is the sensitivity matrix that describes how individual weight elements influence the KPIs, and $\Delta {W} = {W}_{\ell+1} - {W}_\ell$ is the weight increment. The weight update is computed by solving a regularized quadratic program that balances KPI minimization against excessive weight changes:
\begin{equation}
\Delta {W}^\star	= \arg\min_{\Delta {W}}	\left\|	{e}_\ell + {S} \Delta 
{W}	\right\|_{\alpha}^2	+	\left\|	\Delta {W}	\right\|_{{\beta}}^2,
\label{eq:ilc:cost}
\end{equation}
where $\alpha\succeq 0$ prioritizes different KPIs (e.g., assigning higher weight to tracking accuracy than control smoothness) and $\beta \succ 0$ regularizes the update magnitude to ensure smooth adaptation. Because the objective is strictly convex, the optimal update admits a closed-form solution:
\begin{subequations}
\begin{align}
&\Delta {W}^\star= -\left({S}^\top {\alpha} {S} + {\beta}	\right)^{-1}{S}^\top {\alpha} {e}_\ell,	\\
&{W}_{\ell+1} = {W}_\ell + \Delta {W}^\star.
\end{align}
\label{eq:ilc:closedform}
\end{subequations}
This update resembles a single Gauss--Newton iteration on a regularized least-squares objective, providing rapid convergence while maintaining numerical stability. A significant difference from standard NO-ILC methods is that our proposed algorithm does not use gradient-based updates of the NMPC weights. Such an approach would require differentiating the full nonlinear optimal control problem with respect to the matrices $(Q, R)$. The NMPC cost is minimized only implicitly through the solution of a nonlinear program whose states and inputs $({x}_k^{\star},{u}_k^{\star})$ depend on the weights via the KKT conditions, line-search globalization, and solver internals. Computing $\partial J/\partial Q$ or $\partial J/\partial R$ would therefore require implicit differentiation of the entire OCP solution map, including second-order derivatives of the dynamics, constraints, and solver regularization terms. This procedure is computationally expensive, numerically fragile, and incompatible with the fact that the outer-loop objective is not the NMPC cost itself, but task-level KPIs such as tracking RMSE, control smoothness, and saturation ratio.

To circumvent these difficulties, we adopt a sensitivity model as in~\eqref{eq:ilc_affine}, where $S$ captures the empirical influence of weight changes on KPI errors across repetitions. Since the mapping from weight vector $W$ to KPI error $e_\ell$ is nonlinear and implicit---each evaluation of $e_\ell$ requires solving a full closed-loop NMPC simulation---analytic derivatives $\partial e_\ell/\partial W$ are not available. Instead, we construct $S$ empirically: for each weight component, we apply a small perturbation $\delta W$, re-run the closed-loop NMPC, and measure the resulting KPI deviation $\delta e_\ell$, yielding $S \approx \frac{\delta e_\ell}{\delta W}$.

The perturbation magnitudes and matrix structure are determined through empirically to balance approximation accuracy with computational efficiency. Using $S$, the weight update follows a regularized least-squares step analogous to norm-optimal ILC (see~\eqref{eq:ilc:closedform}), where $\beta \succ 0$ is a regularization matrix. This empirical approach avoids differentiating through the NMPC optimizer while providing a practical, robust update law for systems where explicit differentiation is infeasible.

\subsection{Algorithm Summary}
\label{subsec:algo}
Algorithm~\ref{alg:ilc_nmpc} summarizes the complete iterative tuning procedure for NMPC weight tuning. The algorithm initializes with default weights, executes the task under NMPC control, evaluates performance via KPIs, updates the sensitivity matrix, computes new weights using the norm-optimal law~\eqref{eq:ilc:closedform}, and repeats until the weights achieve the desired performance, according to the definition of the corresponding KPIs~\eqref{eq:ek:definition}.

\begin{algorithm}[h!]
\caption{Iterative tuning for NMPC weight tuning.}
\label{alg:ilc_nmpc}
\begin{algorithmic}[1]
\STATE \textbf{Initialize:} ${W}_0$, ${S}$, ${\alpha}$, ${\beta}$, 
$\ell=0$
\WHILE{$\|{e}_\ell\| > \epsilon$ and $\ell < \ell_{\max}$}
\STATE Execute task with NMPC using weights ${W}_\ell$
\STATE Measure output trajectory ${y}_\ell(t)$ and control inputs ${u}_\ell(t)$
\STATE Compute KPI vector ${e}_\ell$ from task performance metrics
\STATE Compute weight update: $\Delta {W}^\star = -({S}^\top {\alpha} 
{S}+ {\beta})^{-1} {S}^\top {\alpha} {e}_\ell$
\STATE Update weights: ${W}_{\ell+1} = {W}_\ell+ \Delta {W}^\star$
\STATE $\ell \leftarrow \ell + 1$
\ENDWHILE
\STATE \textbf{Return:} Tuned weights ${W}^*$
\end{algorithmic}
\end{algorithm}

The regularization term ${\beta}$ in~\eqref{eq:ilc:cost} prevents drastic weight changes that could destabilize the closed-loop system or violate NMPC feasibility. Furthermore, the KPI-based formulation naturally handles trade-offs 	between competing objectives (tracking vs. smoothness vs. actuator usage), achieving multi-objective optimization.
\section{Case Study: Robotic Winding Process}
\label{sec:winding}
Robotic automated fiber placement enables manufacturing of complex composite structures beyond the capabilities of traditional filament winding machines~\citep{dobrich2025planar}. However, achieving consistent winding quality across non-axisymmetric geometries demands precise trajectory generation and control, uniform fiber tension, and proper surface contact—requirements that are highly sensitive to controller parameters and geometric variations.

This work develops an autonomous control framework with three key objectives: (1) generate collision-free winding trajectories satisfying geometric and process constraints, (2) achieve precise end-effector tracking with smooth control actions, and (3) automatically adapt controller parameters across geometries without manual tuning. We address these through a hierarchical architecture integrating model predictive control for planning and execution with iterative tuning for parameter refinement.

The proposed system integrates five components: (1) winding core geometry specification defining the tetrahedral target structure, (2) NMPC-based path planner generating kinematically feasible winding trajectories, (3) closed-loop NMPC controller tracking references in real-time with weights $Q$ and $R$, (4) \ur~robot plant executing controls and providing state feedback, and (5) iterative tuning module adapting NMPC weights across repetitions based on KPI performance. This architecture separates geometric planning, real-time trajectory tracking, and iterative performance optimization into distinct functional layers. The following subsections detail each component's implementation.
\subsection{\ur~Robot Modeling}
\label{subsec:modeling}
\subsubsection{Forward and Inverse Kinematics:}
\label{subsubsec:fkik}
The \ur~collaborative robot is a 6-Degree of Freedom (DOF) serial manipulator with a spherical wrist configuration. The forward kinematics map joint angles ${q} = [q_1, q_2, \ldots, q_6]^\top$ to the end-effector pose ${P_{\rm EE}} = [{p}^\top, {\phi}^\top]^\top$, where ${p} \in \mathbb{R}^3$ denotes Cartesian position and ${\phi} \in \mathbb{R}^3$ represents orientation (e.g., ZYZ Euler angles). Using the Denavit-Hartenberg (DH) convention with calibrated parameters specific to our \ur~unit, the homogeneous transformation from base to end-effector is:
\begin{equation}
{T}_6^0({q}) = \prod_{i=1}^{6} {A}_j(q_j),
\label{eq:fk:transform}
\end{equation}
where ${A}_j(q_j)$ is the transformation matrix for joint $j$ based on its DH parameters~\citep{ur2025dh} (for more details on forward kinematic refer to~\cite{hawkins2013analytic, petrone2025dynamic}).

The inverse kinematics problem admits multiple solutions due to the spherical wrist. We employ an analytical inverse kinematics solver presented in~\cite{hawkins2013analytic} that exploits the wrist decoupling property to compute all feasible joint configurations for a desired end-effector pose ($P_{\rm EE}$). Among the candidate solutions, we select the configuration that minimizes the weighted joint displacement from the current state while respecting joint limits.
\subsubsection{Dynamic Model:}
\label{subsubsec:dynamic}
The robot dynamics are described by the manipulator equation:
\begin{equation}
{M}({q})\ddot{{q}} + {C}({q}, \dot{{q}})\dot{{q}} + {G}({q}) = {\tau},
\label{eq:robot_dynamics}
\end{equation}
where ${M}({q}) \in \mathbb{R}^{6 \times 6}$ is the inertia matrix, ${C}({q}, \dot{{q}}) \in \mathbb{R}^{6 \times 6}$ represents Coriolis and centrifugal effects, ${G}({q}) \in \mathbb{R}^6$ captures gravitational torques, and ${\tau} \in \mathbb{R}^6$ is the joint torque vector.

For NMPC formulation, we adopt a state-space representation with joint positions as states ($x$) and joint velocities as control inputs ($u$):
\begin{equation}
\begin{aligned}
{x} &= [{q}^\top, {\tau}^\top]^\top \in \mathbb{R}^{12}, \\
{u} &= \dot{{q}} \in \mathbb{R}^6, \\
{y} &= {h}({q}) = [{p}({q})^\top, {\phi}({q})^\top]^\top \in \mathbb{R}^6,
\end{aligned}
\label{eq:model:robot}
\end{equation}
where ${y}$ represents the end-effector pose and orientation computed via forward kinematics.

%

\subsection{Open-Loop NMPC-based Path Planner}
\label{subsec:pp}
The path planner generates a dynamically feasible end-effector trajectory that follows a geometric path defined by the core vertices $(v_m)$, their segment tangents, the Tool Center Point (TCP) thread-contact offset, and the payout-eye pose $p^{\mathrm{eye}}$. A dense reference path $(p_k^{\mathrm{pp}})$ is obtained by linear interpolation between vertices, with optional corner refinement (increased sampling around sharp direction changes) and tapered vertex weights for improved accuracy at geometric corners.

Given the initial robot state $x_0$, previous inputs $u_{-1}$, and the interpolated reference $(p_k^{\mathrm{pp}})$, the planner solves the finite-horizon open-loop NMPC problem:
\begin{subequations}
\label{eq:pp:nmpc}	
\begin{flalign}
\label{eq:pp:nmpc:cost}
&\min_{U}\;
\sum_{k=0}^{N-1} \Big(
\underbrace{\|p_k - p_k^{\mathrm{pp}}\|_{\tilde{Q}}^2
+ w_v \|p_k - v_k\|^2}_{\text{position + vertex tracking}}  + \underbrace{|u_k -  u_{k-1}\|_{\tilde{R}}^2}_{\text{smoothness}}
&& \nonumber \\
&\qquad\qquad
+\underbrace{w_{\mathrm{align}}\;\Psi_{\mathrm{thread}}(p_{c,k}, \mathbb{T}_k)}_{\text{thread alignment}}
\Big)
\end{flalign}
\vspace{-6mm}
\begin{align}
\nonumber
&\text{s.t.}\\[1mm]
\label{eq:pp:nmpc:model:states}
&\text{robot dynamics in}~\eqref{eq:model:robot},\\
\label{eq:pp:nmpc:const:states}
& q_{\min} \le q_k \le q_{\max},\\
\label{eq:pp:nmpc:const:input}
& \dot{q}_{\min} \le \dot{q}_k \le \dot{q}_{\max},\\
& d_{\mathrm{obs}}(p_k) \ge d_{\min},\quad\text{(workspace clearance)},\\
& d_{\mathrm{self}}(q_k) \ge d_{\min}^{\mathrm{self}},\quad\text{(self-collision capsules)},\\
& k=0,\dots,N-1.
\end{align}
\end{subequations}
Here, $u_k$ are joint velocities, $p_k$ is the predicted end-effector position, $v_k$ are interpolated vertex points, $p_{c,k}$ is the TCP thread-contact point, $\mathbb{T}_k$ is the local groove tangent, and $w_{\mathrm{thread}}$, $w_{\mathrm{ori}}$ denote smooth thread-alignment and orientation penalties. Self-collision distances $d_{\mathrm{self}}(q_k)$ are computed using capsule models of the robot links, following standard swept-sphere geometry~\citep{zimmermann2022differentiable}. The planner returns the optimal open-loop sequences
$(u_k^\star)$, $(x_k^\star)$, and $(p_k^\star)$, which are used as reference trajectories for closed-loop NMPC tracking. It is worth to mention that the weights in the~\eqref{eq:pp:nmpc:cost} were fixed throughout the planning. 

\subsection{Closed-Loop NMPC Trajectory Tracking Controller}
\label{subsec:cl-nmpc}
The path planner provides a time-parameterized end-effector position trajectory $\{p_k^{\star}\}_{t=0}^{T_{f}}$ in the robot base frame. At each control instant $t$, with measured state ${x}(t)$ and previous input $u(t-T_s)$, the tracking NMPC computes joint-velocity commands that follow this trajectory while respecting joint and input limits.

Let $p_{k}^{\mathrm{ref}} = p_{k}^{\star}$ denote the reference position supplied to the controller over its horizon. Given the initial robot state $x_0$, previous inputs $u_{-1}$, the closed-loop NMPC at time $t$ solves:
\begin{subequations}
\label{eq:tc:nmpc}
\begin{align}
\label{eq:tc:nmpc:cost}
&\min_{U}\;\sum_{k=0}^{N-1} \Big(\underbrace{\|p_k- p_{k}^{\mathrm{ref}}\|_{Q}^2}_{\text{position tracking}}  + \underbrace{|u_k - u_{k-1}\|_{{R}}^2}_{\text{smoothness}}\Big)
\end{align}
\begin{align}
\nonumber
&\text{s.t.}\\
\label{eq:tc:nmpc:model:states}
&\text{robot dynamics in}~\eqref{eq:model:robot},\\
\label{eq:tc:nmpc:const:states}
& q_{\min} \le q_k \le q_{\max},\\
\label{eq:tc:nmpc:vel-limits}
& \dot{q}_{\min} \le u_k \le \dot{q}_{\max}, \\
\label{eq:tc:nmpc:torque-limits}
& \tau_{\min} \le \tau_k \le \tau_{\max}\\
& k=0,\dots,N-1.
\end{align}
\end{subequations}
Here, $u_k$ are joint velocities, $p_k$ is the predicted end-effector position, and $p_{k}^{\mathrm{ref}}$ is the corresponding reference position from the path planner. The matrices $Q\succ 0$ and $R\succ 0$ are weights on the position tracking and input effort, respectively. 
\section{Simulation Results}
\label{sec:results}
To evaluate the performance of the proposed iterative tuning NMPC framework, we conduct extensive simulations of the robotic carbon fiber winding process on a tetrahedral core (edge length 80~mm). The reference winding trajectories are generated using the path planning algorithm described in Section~\ref{subsec:pp}. In the path planner, we used a fixed sampling time $T_s = 8$~ms and a prediction horizon of $N = 10$ samples with $\bar{Q} = \mathbb{I}_{3\times3}*2 \times 10^4$, $\bar{R} = \mathbb{I}_{6\times6}*5\times 10^{-3}$, and $w_{\text{align}} = 1\times 10^{3}$. For the trajectory control, we compare two control approaches:
\begin{itemize}
\item \textbf{Fixed-weight NMPC (Controller 1):} it employs standard NMPC with fixed weight matrices determined through offline Bayesian Optimization. BO is applied to tune the NMPC weights $Q$ and $R$ for the carbon fiber winding task on a tetrahedral core, targeting the KPI objectives defined in~\eqref{eq:ek:definition}. The optimization is performed using MATLAB's \texttt{bayesopt} function~\citep{matlab2025bayesopt} configured with an \texttt{ARD Matérn 5/2} kernel for the Gaussian process surrogate model and its parameters are optimized via negative log-likelihood minimization, and the \texttt{expected-improvement-plus} acquisition function to balance exploration and exploitation. The maximum number of objective evaluations is set to 100 to limit computational cost while ensuring convergence. Once the optimal weights are identified, they remain fixed throughout all repetitions of the winding task.
\item \textbf{NMPC with iterative weights tuning (Controller 2):} proposed iterative tuning NMPC as described in Section~\ref{subsec:noilc}.
\end{itemize}

Both controllers are implemented in MATLAB with CasADi toolbox and IPOPT solver for optimization~\citep{andersson2019casadi}. The NMPC formulations use a prediction horizon of $N = 10$ steps with a sampling time of $T_s = 8$~ms. The bounds on the weights were set to $Q = [1 \;\;10^6]$ and $R = [10^{-6}\;\; 1]$. The simulation comprises $\ell_{\max} = 10$ winding repetitions for the tetrahedral core. For the iterative tuning NMPC approach, the weighting matrices are initialized with identity matrix (see Table~\ref{tab:q:values} and~\ref{tab:r:values}). For the iterative tuning NMPC approach (Controller 2), the weighting matrices are initialized with identity matrices ($Q = \mathbb{I}_{3 \times 3}$, $R = \mathbb{I}_{6 \times 6}$) at the start of the first repetition, as shown in Tables~\ref{tab:q:values} and~\ref{tab:r:values}. After each winding repetition, the algorithm evaluates task-level KPIs and updates the weights according to the iterative tuning law~\eqref{eq:ilc:closedform}. This adaptation process continues across repetitions until the measured performance satisfies the target specifications (RMSE $\leq$ 1~mm, max error $\leq$ 2~mm, RMS control $\leq$ 0.025~rad/s). Once convergence is achieved, the weights are held constant for all subsequent repetitions. The results and weight values presented in this section correspond to the final converged state after 4 iterations.

\begin{figure}[h!]
\centering
\includegraphics[scale=0.36]{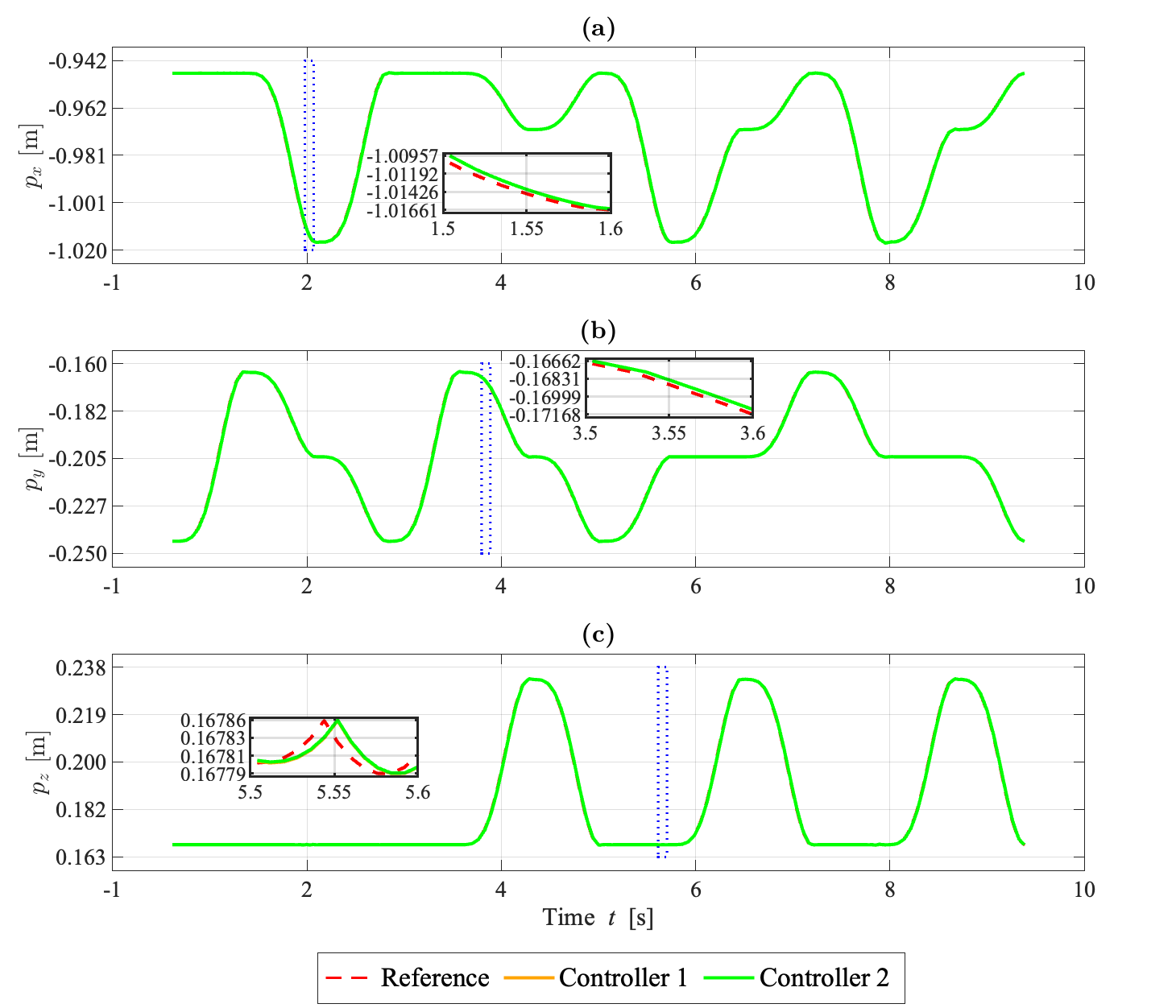}
\caption{End-effector position tracking for X, Y, and Z coordinates of the \ur~manipulator for Controller 1 and 2.}
\label{fig:ee_tracking}
\end{figure}

Fig.~\ref{fig:ee_tracking} presents the end-effector position tracking in Cartesian coordinates ($p_x$, $p_y$, $p_z$) for both controllers over the complete winding trajectory. The reference trajectory, obtained from the geometric path planner for the tetrahedral core, exhibits varying dynamics across different segments of the task. Both Controller 1 (fixed weights from Bayesian Optimization) and Controller 2 (online adaptive weights) track the reference closely throughout the entire trajectory. Visual inspection reveals negligible differences between the two controllers, which is quantitatively confirmed by the performance metrics in Table~\ref{tab:performance:metrics}: Controller 2 achieves RMSE within 0.30\% and maximum error within 0.30\% of Controller 1. The zoomed insets highlight regions where the trajectory undergoes rapid changes, demonstrating that both controllers maintain tight tracking even during demanding maneuvers.

\begin{figure}[h!]
\centering
\includegraphics[scale=0.35]{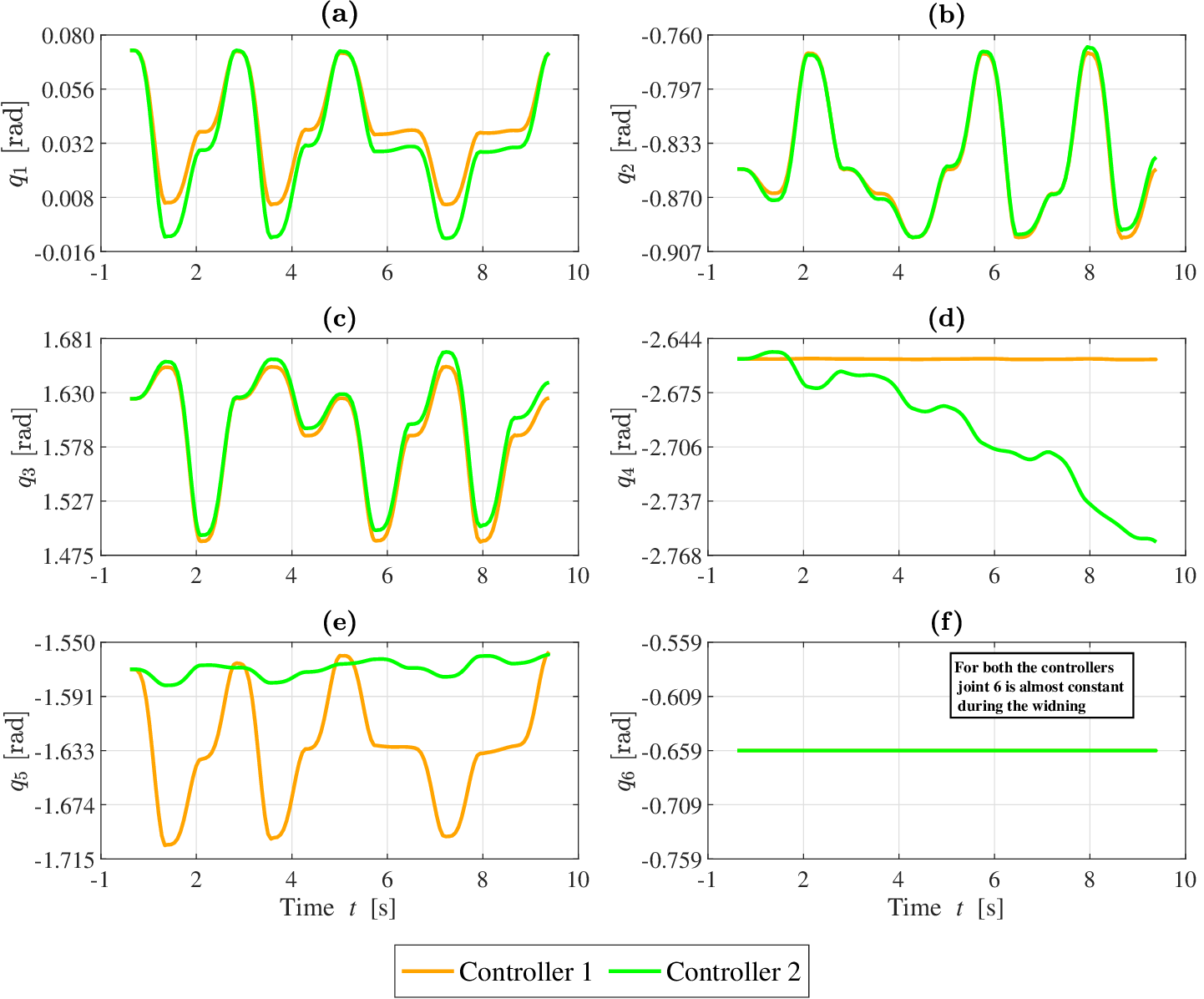}
\caption{Joint position trajectories for all six joints of the \ur~manipulator for Controller 1 and 2.}
\label{fig:joint_positions}
\end{figure}

\begin{figure}[h!]
\centering
\includegraphics[scale=0.35]{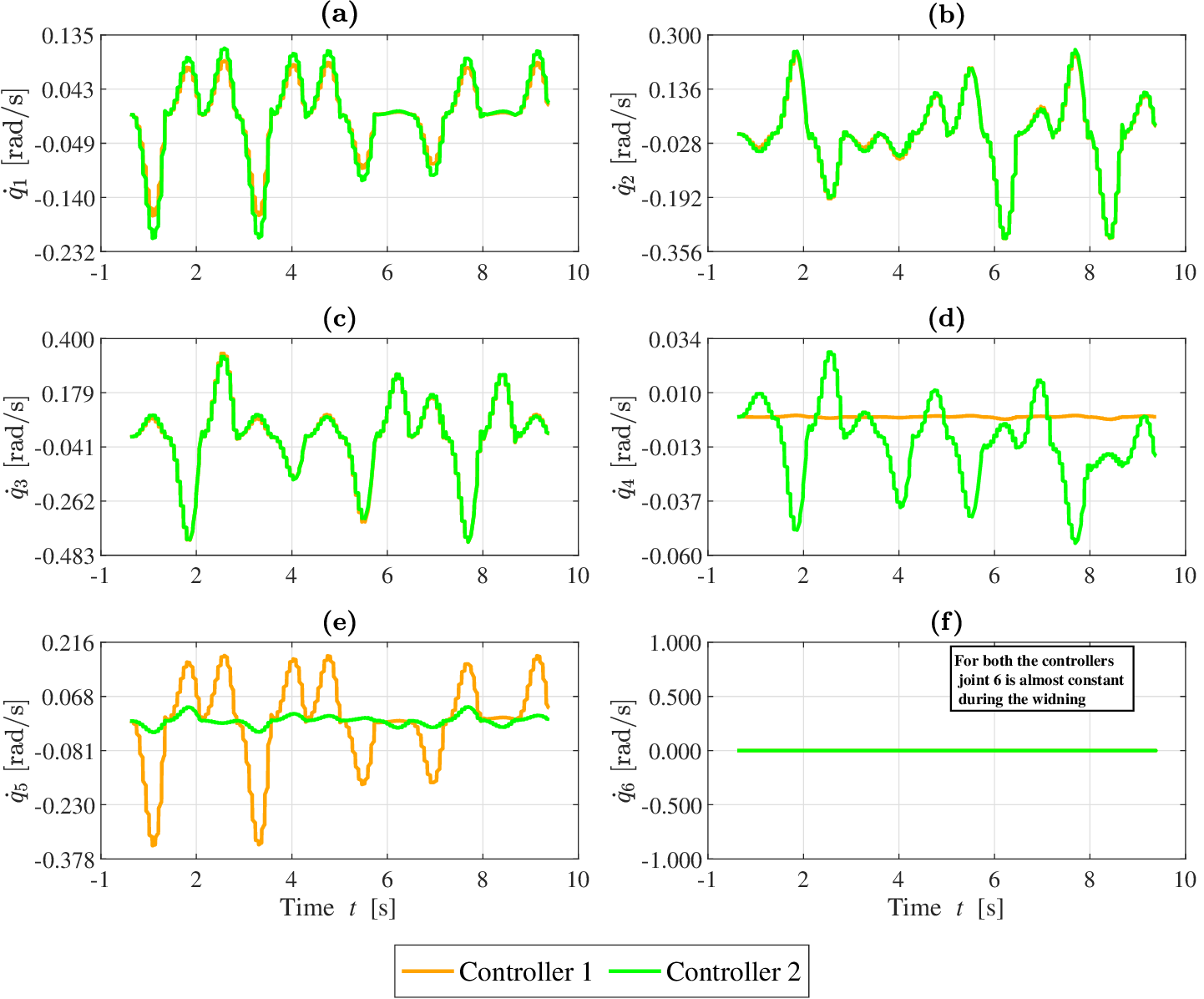}
\caption{Joint velocity profiles showing control actions of the \ur~manipulator for Controller 1 and 2.}
\label{fig:joint_velocities}
\end{figure}

Joint-level trajectories are illustrated in Fig.~\ref{fig:joint_positions} and~\ref{fig:joint_velocities}. Fig.~\ref{fig:joint_positions} compares the six joint angle profiles for both controllers throughout the winding cycle. Several key observations emerge from this comparison. First, both controllers successfully respect joint position constraints throughout the entire trajectory. Second, joint 6 remains stationary for both controllers, consistent with the constant reference provided by the path planner to maintain tool orientation. Third, joints 1-5 demonstrate smooth tracking under both control strategies, with Controller 2 achieving joint-space performance nearly indistinguishable from Controller 1. The most dynamic joints are 3 and 5, which govern large-scale end-effector positioning and exhibit the largest angular excursions. Even for these demanding joints, Controller 2 matches the tracking quality of Controller 1, with differences barely perceptible. This joint-level analysis reinforces the end-effector tracking results (Fig.~\ref{fig:ee_tracking}), confirming that online iterative weight tuning achieves control performance equivalent to extensive offline optimization.

Fig.~\ref{fig:joint_velocities} reveals critical insights into control smoothness through joint velocity profiles. A direct comparison shows that both controllers produce remarkably similar velocity characteristics. Controller 1 demonstrates smooth, well-regulated velocities with gradual accelerations and minimal oscillatory content. Controller 2 matches this smoothness, and in fact achieves slightly superior performance with 13\% lower RMS control velocity (Table~\ref{tab:performance:metrics}). Both controllers exhibit the most significant velocity variations in joints 4-5, corresponding to their role in large-scale end-effector positioning, while maintaining smooth profiles without excessive fluctuations. Joint 6 remains at zero velocity, consistent with its constant position. Neither controller shows problematic characteristics such as frequent velocity reversals, high-frequency oscillations, or abrupt changes that would increase energy consumption or mechanical stress. The comparable—and in some metrics superior—velocity smoothness of Controller 2 demonstrates that online weight adaptation can match the control quality achieved through extensive offline optimization.

\subsection{Quantitative Performance Comparison}
\label{subsec:comp}
Table~\ref{tab:performance:metrics} presents a comprehensive comparison of the two control strategies across learning performance, tracking accuracy, and control characteristics.

\subsubsection{Learning Performance}
Controller 1 requires 100 offline Bayesian Optimization evaluations to determine optimal weights before task execution. In contrast, Controller 2 converges to comparable performance in just 4 online repetitions during task repetitions, demonstrating significantly more efficient weight adaptation—a 25x reduction in tuning evaluations.

\subsubsection{Tracking Accuracy}
Both controllers achieve excellent tracking performance that meets the target specifications (RMSE $\leq$ 1~mm, maximum error $\leq$ 2~mm). Controller 1 attains RMSE of 0.990~mm, maximum error of 1.645~mm, and mean error of 0.800~mm. Controller 2 exhibits nearly identical tracking with RMSE of 0.993~mm, maximum error of 1.650~mm, and mean error of 0.802~mm—representing differences of merely 0.30\%, 0.30\%, and 0.25\%, respectively. These minimal deviations confirm that online iterative weight tuning successfully replicates the precision of extensive offline optimization.

\subsubsection{Control Characteristics}
Controller 1 achieves lower total control effort at 298.142~rad/s, while Controller 2 requires 9.74\% more at 327.189~rad/s. However, Controller 2 demonstrates superior control smoothness with RMS control of 0.020~rad/s compared to Controller 1's 0.023~rad/s—a 13.04\% improvement indicating smoother control actions that reduce actuator wear and mechanical stress. Computational efficiency also favors Controller 2, which executes 14.98\% faster at 2.757~ms per repetition compared to Controller 1's 3.170~ms. Computation times are the average times calculated for online optimization with IPOPT solver. Both computation times easily satisfy real-time control requirements for the robotic system.

\subsubsection{Overall Assessment}
Controller 2 achieves tracking accuracy within 0.3\% of Controller 1 while offering several practical advantages: (1) elimination of 100 offline optimization evaluations through convergence in just 4 online repetitions, (2) elimination of offline computation time for Bayesian Optimization, (3) 15\% faster online computation time enabling more responsive control, (4) 13\% smoother control actions reducing mechanical stress, and (5) online adaptability to task variations or disturbances. These benefits come at the cost of marginally higher control effort (9.74\% increase), which remains well within the system's capabilities and represents a favorable trade-off for iterative robotic applications requiring both precision and adaptability.

\begin{table}[h]
\centering
\setlength{\tabcolsep}{2pt} 
\renewcommand*{\arraystretch}{0.9}
\caption{Performance metrics for the robotic winding controllers (10 winding repetitions).}
\label{tab:performance:metrics}
\begin{tabular}{lccc}
\toprule
\textbf{KPI Metric} & \textbf{Ctrl. 1} & \textbf{Ctrl. 2} & \textbf{Change [\%]} \\
\midrule
\multicolumn{4}{l}{\textit{Learning Performance}} \\
Convergence repetition & BO Offline (100) & \textbf{4} & +96 \\
\midrule
\multicolumn{4}{l}{\textit{Tracking Accuracy}} \\
RMSE [mm] & \textbf{0.990} & 0.993 & $-0.30$ \\
\addlinespace
Max error [mm] & \textbf{1.645} & 1.650 & $-0.30$ \\
\addlinespace
Mean error [mm] & \textbf{0.800} & 0.802 & $-0.25$ \\
\midrule
\multicolumn{4}{l}{\textit{Control Characteristics}} \\
Control effort [rad/s] & \textbf{298.142} & 327.189 & $+9.74$ \\
\addlinespace
RMS Control [rad/s] & \textbf{0.023} & 0.020 & $+13.04$ \\
\addlinespace
Computation time [ms] & 3.170 & \textbf{2.757} & $+14.98$ \\
\bottomrule
\end{tabular}
\begin{tablenotes}
\small
\item $^\star$Percentages show relative change from Fixed NMPC (Controller 1)
\item $^\star$Bold indicates best performance for each metric
\end{tablenotes}
\end{table}

Tables~\ref{tab:q:values} and~\ref{tab:r:values} present the evolution of NMPC weight matrices across repetitions for both controllers. Controller 1 employs fixed weights ($Q_{1,1} = 9.606 \times 10^5$, $Q_{2,2} = 1.711 \times 10^5$, $Q_{3,3} = 9.988 \times 10^5$) optimized offline via Bayesian Optimization, which remain constant throughout all repetitions. In contrast, Controller 2 begins with uniform initial weights ($Q_{i,i} = 1$, $R_{i,i} = 1$) and adapts them online after each repetition. The $Q$ weights converge from initial values to approximately 334 by repetition 4, while $R$ weights decrease significantly from 1.0 to $10^{-5}$, reflecting the controller's learning to prioritize tracking accuracy while reducing control effort. Once convergence is achieved at repetition 4, both controllers maintain their respective weights for subsequent repetitions.
\begin{table}[h]
\centering
\caption{Output ($Q$) weight values across repetitions for both controllers.}
\label{tab:q:values}
\begin{tabular}{l|c|cccc}
\toprule
& \textbf{Ctrl. 1} & \multicolumn{4}{c}{\textbf{Ctrl. 2}} \\
\midrule
\textbf{Weight} $\downarrow$ & \multicolumn{5}{c}{\textbf{Repetition} $\rightarrow$} \\
& \textbf{1} & \textbf{1} & \textbf{2} & \textbf{3} & \textbf{4} \\
\midrule
$Q_{1,1}$ & $9.606 \times 10^{5}$ & 1 & 317.641 & 327.257 & 334.175 \\
$Q_{2,2}$ & $1.711 \times 10^{5}$ & 1 & 317.641 & 327.257 & 334.175 \\
$Q_{3,3}$ & $9.988 \times 10^{5}$ & 1 & 317.641 & 327.257 & 334.175 \\
\bottomrule
\end{tabular}
\end{table}
\begin{table}[h]
\centering
\setlength{\tabcolsep}{3pt} 
\caption{Input ($R$) weight and control effort values across repetitions for both controllers.}
\label{tab:r:values}
\begin{tabular}{l|c|cccc}
\toprule
& \textbf{Ctrl. 1} & \multicolumn{4}{c}{\textbf{Ctrl. 2}} \\\midrule
\textbf{Weight} $\downarrow$ & \multicolumn{5}{c}{\textbf{Repetition} $\rightarrow$} \\
& \textbf{1} & \textbf{1} & \textbf{2} & \textbf{3} & \textbf{4} \\
\midrule
$R_{1,1}$ & 0.029 & 1 & 0.212 & 0.097 & $10^{-5}$ \\
$R_{2,2}$ & $3.9 \times 10^{-4}$ & 1 & 0.212 & 0.097 & $10^{-5}$ \\
$R_{3,3}$ & $1.7 \times 10^{-5}$ & 1& 0.212 & 0.097 & $10^{-5}$ \\
$R_{4,4}$ & 0.779 & 1 & 0.212 & 0.097 & $10^{-5}$ \\
$R_{5,5}$ & 0.177 & 1 & 0.212 & 0.097 & $10^{-5}$ \\
$R_{6,6}$ & $2.5 \times 10^{-5}$ & 1 & 0.212 & 0.097 & $10^{-5}$ \\
Control Efforts & 305.59 & 814.79 & 336.36 & 334.93 & 327.48 \\
\bottomrule
\end{tabular}
\end{table}
\section{Conclusions}
\label{sec:conclusion}
This paper presented an iterative learning framework for online NMPC weight tuning in repetitive robotic tasks. Using an empirical sensitivity matrix and norm-optimal ILC principles, the method automatically adapts weights $Q$ and $R$ based on KPI feedback across repetitions. Validation on simulated carbon fiber winding demonstrated convergence to tracking accuracy within 0.3\% of Bayesian Optimized NMPC performance in 4 repetitions versus 100 offline evaluations, with additional advantages in control smoothness (13\% improvement) and computation time (15\% faster). The framework provides a practical alternative to offline optimization for NMPC tuning, eliminating pre-deployment tuning overhead while enabling online adaptation. Future work will focus on experimental validation, time-varying tasks, and analysis of the convergence of the algorithm.
\begin{ack}
Authors would like to thank Dr. Gabriele Fadini and Sarah Steiner for their valuable suggestions and helpful discussions.
\end{ack}
\bibliography{ifacconf} 
\end{document}